\documentclass[10pt,twocolumn,letterpaper]{article}

\usepackage[pagenumbers]{cvpr} % To force page numbers, e.g. for an arXiv version

\usepackage{graphicx}
\usepackage{amsmath}
\usepackage{amssymb}
\usepackage{booktabs}
\usepackage{pifont}
\newcommand{\cmark}{\ding{51}}
\newcommand{\xmark}{\ding{55}}

\usepackage{times}
\usepackage{epsfig}
\usepackage{epstopdf}
\usepackage[linesnumbered,ruled,vlined]{algorithm2e}

\SetCommentSty{mycommfont}

\SetKwInput{KwInput}{Input}                % Set the Input
\SetKwInput{KwOutput}{Output}              % set the Output

\usepackage{caption}
\usepackage{blindtext}
\usepackage{url}
\usepackage{multirow}

\usepackage[pagebackref,breaklinks,colorlinks]{hyperref}

\usepackage[capitalize]{cleveref}
\crefname{section}{Sec.}{Secs.}
\Crefname{section}{Section}{Sections}
\Crefname{table}{Table}{Tables}
\crefname{table}{Tab.}{Tabs.}

\def\cvprPaperID{4230} % *** Enter the CVPR Paper ID here
\def\confName{CVPR}
\def\confYear{2022}

\begin{document}

\title{Freeform Body Motion Generation from Speech}

\author{Jing Xu\textsuperscript{1}~~~~~~~~Wei Zhang\textsuperscript{2}~~~~~~~Yalong Bai\textsuperscript{2}~~~~~~~~Qibin Sun\textsuperscript{1}~~~~~~~~Tao Mei\textsuperscript{2}\\
{\normalsize \textsuperscript{1}University of Science and Technology of China~~~~~~~~\textsuperscript{2}JD AI Research}\\
{\tt\small xujing0@mail.ustc.edu.cn, wzhang.cu@gmail.com, ylbai@outlook.com,} 
\\
{\tt\small qibinsun@ustc.edu.cn,  tmei@live.com}
}

\maketitle

%%%%%%%%% ABSTRACT
\begin{abstract}
People naturally conduct spontaneous body motions to enhance their speeches while giving talks.
Body motion generation from speech is inherently difficult due to the non-deterministic mapping from speech to body motions. 
Most existing works map speech to motion in a deterministic way by conditioning on certain styles, leading to sub-optimal results.
Motivated by studies in linguistics, we decompose the co-speech motion into two complementary parts: pose modes and rhythmic dynamics.
Accordingly, we introduce a novel freeform motion generation model (FreeMo) by equipping a two-stream architecture, i.e., a pose mode branch for primary posture generation, and a rhythmic motion branch for rhythmic dynamics synthesis. On one hand, diverse pose modes are generated by conditional sampling in a latent space, guided by speech semantics. On the other hand, rhythmic dynamics are synced with the speech prosody. Extensive experiments demonstrate the superior performance against several baselines, in terms of motion diversity, quality and syncing with speech.
Code and pre-trained models will be publicly available through \href{https://github.com/TheTempAccount/Co-Speech-Motion-Generation}{URL}.
\end{abstract}

%%%%%%%%% BODY TEXT
\section{Introduction}

Body motion generation from speech is to synthesize spontaneous body motions synchronized with input speech audio.
Professional speakers are experts in utilizing such motions to effectively deliver information. 
This task is essential for applications such as digital avatars and social robots. 
Notably with this technique, amateur speakers can also generate their own ``professional'' talking videos, by mimicking moves from professional speakers.

While generating lip motions has been extensively studied in talking heads generation \cite{talking_head_survey}, synthesizing plausible co-speech body motions remains an open issue.
Specifically, lip motions can be well matched with the input audio using a deterministic mapping, i.e., one to one mapping from phonemes\footnote{Basic elements of speech audio.} to lip shapes. However, such models can not be trivially extended to body, due to the highly stochastic nature of body motions during a talk speech. 
Practically, the co-speech body motion is highly \emph{freeform}. Even if the same person gives the same speech twice in a row, there is no guarantee that the speaker would exhibit the same body motions. Moreover, a person usually switches poses from time to time during a long talking speech. 
As in Fig.~\ref{fig:diversity}, the same speech audio does not necessarily lead to a fixed form of motions, and different speeches may go well with the same motion sequence. 
% The non-deterministic mapping from speech to body motion is neglected in prior studies. 

Most existing works treat body and lip motion generation in a similar way, i.e., the body landmarks are directly inferred from the input audio via a deep network  \cite{Audio-driven_talking_face_head_pose, talking_head_generation_with_rhythmic_head_motion, speech2video_synthesis, speech2gesture, trimodal, speech_drives_templates, style_transfer}. 
To simplify the non-deterministic mapping, some methods rely on a set of pre-defined gestures \cite{speech2video_synthesis}, or condition on person-specific styles \cite{speech2gesture, trimodal} and templates \cite{speech_drives_templates}.
%For example, \cite{speech2video_synthesis} relies on a set of pre-defined gestures to bypass the complex mapping, and \cite{speech2gesture, trimodal}  simplify the complex mapping by conditioning on person-specific styles. 
These solutions can mimic motions of certain speakers/styles to some degree, but they are limited in terms of motion diversity and fidelity, especially for long talk speeches. Therefore, it is critical to develop algorithms that model the non-deterministic mapping between speech and body motions.

% \begin{figure}[t]
%   \centering
%   \includegraphics[width=0.99\linewidth]{figures/fig1.pdf}
%   \caption{The structure of co-speech motions.}
%   \label{fig:components}
% \end{figure}

% Motivated by studies in linguistics and psychology \cite{gesture_and_speech_in_interaction_an_overview}, co-speech motion helps in the organization and presentation of speech delivery, and contributes in both semantics and intonation. Semantically, the switch points of body posture chunk a long speech into turns or information packages \cite{gesture-visible_action_as_utterance}, such that the structure of the speech is easier to understand. For example, gesture changes accompanying topic shifts better guides the audience into new topics \cite{gesture_and_speech_in_interaction_an_overview}. In terms of intonation, the rhythmic hand movements that match to the prosody of audio could attract attentions of the audience, with the stressed syllable during speech. Moreover, proper rhythmic hand motions also reflects the progress of the speech and deliver a vivid listening experience \cite{hand_movements}. Such co-speech motion usually has no specific linguistic meaning, and varies across speakers \cite{gesture_and_thought}. 

Based on studies in linguistics and psychology, co-speech motion helps the organization and presentation during speech delivery, and contributes in both semantics and intonation. Semantically, body motions guide the discourse organization (e.g., indicating topics and paragraphs), or contribute to the utterance content (e.g., describing shape using hands) \cite{gesture-visible_action_as_utterance}. For example, gesture change accompanying topic shifts better guide the audience into new topics \cite{gesture_and_speech_in_interaction_an_overview}. Besides, some gestures are conventionalized and attached with certain linguistic properties (e.g., ``thumbs up"). These gestures are widely used to facilitate communication. In terms of intonation, the rhythmic movement that matches to the prosody of audio could attract attentions of the audience, with the stressed syllable during speech. Moreover, proper rhythmic motions\footnote{Also called beat gestures in some literature.} also reflect the progress of the speech and deliver a vivid listening experience \cite{hand_movements, Hand_and_Mind}. Such co-speech motion usually has no specific linguistic meaning, and manifests as simple and fast hand dynamics related to the prosody \cite{gesture_and_thought}.

% In practice, professional speakers usually put their bodies to some habitual postures to comfortably give speech. 

Motivated by these studies, we consider the structure of co-speech motions in a novel perspective. We introduce the concept of \emph{pose mode} as the mode of the pose distribution that speakers has for fragments of speech. Considering the speaker's posture in a speech video as a random vector, it follows a multi-modal distribution in the high dimensional space. Modes in such distribution (values with local maximal density) corresponds to conventionalized gestures or the habitual postures of speakers. %Postures during talks are largely determined by personal habits \cite{Hand_and_Mind}, and co-speech motions are found to change with the types of communicative situation \cite{gesturing_on_the_telephone}. 
Our work focuses on motions in talk videos, where speakers organize a long speech around a certain topic. Under this setting, the pose modes are mostly habitual postures with no specific global meaning. Consequently, the structure of co-speech motions can be considered as the sequential transitions of \textit{pose modes} with \textit{rhythmic dynamics} under each pose mode. Therefore, the non-deterministic mapping from speech to body motion is decomposed into two parts: a stochastic mapping from speech semantics to pose modes, and the mapping from speech prosody to rhythmic motion dynamics. 
Our contributions are summarized as follows:
\begin{itemize}
%     %\item We present a large dataset of in-the-wild speech videos containing clean spontaneous co-speech motions.
    %\item We address the non-deterministic mapping from speech to body motions with a 
    
    \item To address the non-deterministic mapping from speech to body motions, we propose to decompose the motion into pose modes and rhythmic motions. The former is stochastically generated with conditional sampling in a VAE latent space, and the latter is effectively inferred by speech prosody. 
    
    % \item We propose to learn The non-deterministic mapping between speech to body motion, Such decomposition, we propose to infer the .
    
    % \item We propose a novel co-speech motion generation model that learns the non-deterministic mapping from speech to body motion. By introducing a conditional sampling strategy, the pose  based on speech semantics and novel objective functions that decomposes the dynamics and postures in the motion sequence, 
    
    \item Extensive experiments demonstrate that our model generates plausible freeform motions well synced with the speech, outperforming other baselines in terms of diversity, quality and syncing with clear margin. %The experimental results show that our model significantly outperforms existing methods.
\end{itemize}

\section{Related Work}
\subsection{Relation between Speech and Body Motion}

According to studies in linguistics and 	
gesture \cite{gesture-visible_action_as_utterance}, there are two main functions of co-speech body motions, namely \textit{substantial} and \textit{pragmatic} gestures. The former one contributes to the speech content (e.g., hands are better suited to describe shapes), while the latter one contributes to the situational embedding such as conveying attitudes or agreement or guiding the discourse organization. Besides, beat gestures manifest simple and fast movements of the hands. Rather than directly conveying meaning, they are synchronized with prosodic events in speech and contribute to the perceiving of the speech and function in the sense of audiovisual prosody \cite{Hand_and_Mind}.

%Co-speech body motions is highly adaptive to various situations \cite{gesture_and_speech_in_interaction_an_overview}. Under the non-interactive settings, the co-speech motions become less listener-oriented, and play an important role in speech organizations. This is evidenced by that the gestures were found under both visibility and non-visibility situations, implying that these gestures have a function predominantly facilitating speech production \cite{gesture_and_speech_in_interaction_an_overview}. \cite{why_do_people_gesture_when_they_speak} reported on blind speakers gesturing when talking to other blind listeners. In this paper, we focus on co-speech motions in stand-up talks, in which one speaker give speech to the audience without much interactions (as opposed to conversational settings).

\subsection{Co-speech Motion Generation}
% Previous works on co-speech motion generation can be grouped into rule-based and learning-based methods. 

\noindent\textbf{Rule-based.}
Rule-based methods rely on pre-defined mapping to generate appropriate body motions. Most approaches rely on a fixed set of gesture templates, and operate on text inputs \cite{gesture_controller}. A typical solution is to associate each word to a gesture \cite{BEAT}. Further improvements include assigning synchronization points \cite{Greta}, or analysing communicative goals \cite{synthesizing_multimodal_utterances}.

\noindent\textbf{Learning-based.}
While learning-based methods are successful on lip motion generation \cite{synthesizing_obama, a_lip_sync_expert, text_based_editing, Audio-driven_talking_face_head_pose, talking_head_generation_with_rhythmic_head_motion}, the application on co-speech motion generation is far from mature. Previous methods typically treat co-speech body and lip motions in a similar way. However, talking head generation \cite{Audio-driven_talking_face_head_pose, talking_head_generation_with_rhythmic_head_motion} mostly predicts facial motions in a deterministic mapping. Recent methods model co-speech motion generation as a sequence-to-sequence translation problem. For example, recurrent neural networks \cite{robots_learn_social_skills} are adopted to generate motions from a speech transcript. Speech2gesture \cite{speech2gesture} uses 1D fully-convolutional networks to map speech to motion sequences. Trimodal-context \cite{trimodal} takes both text and audio as input, and synthesizes body motions using recurrent networks. However, these methods learn a deterministic mapping from speech to motion, while essentially this mapping is non-deterministic. 
% In contrast with these methods, we aims to learn the non-deterministic mapping from speech to body motion. 

\section{Freeform Co-speech Motion Generation}
\label{sec: method}

\begin{figure*}
  \centering
  \includegraphics[width=0.89\linewidth]{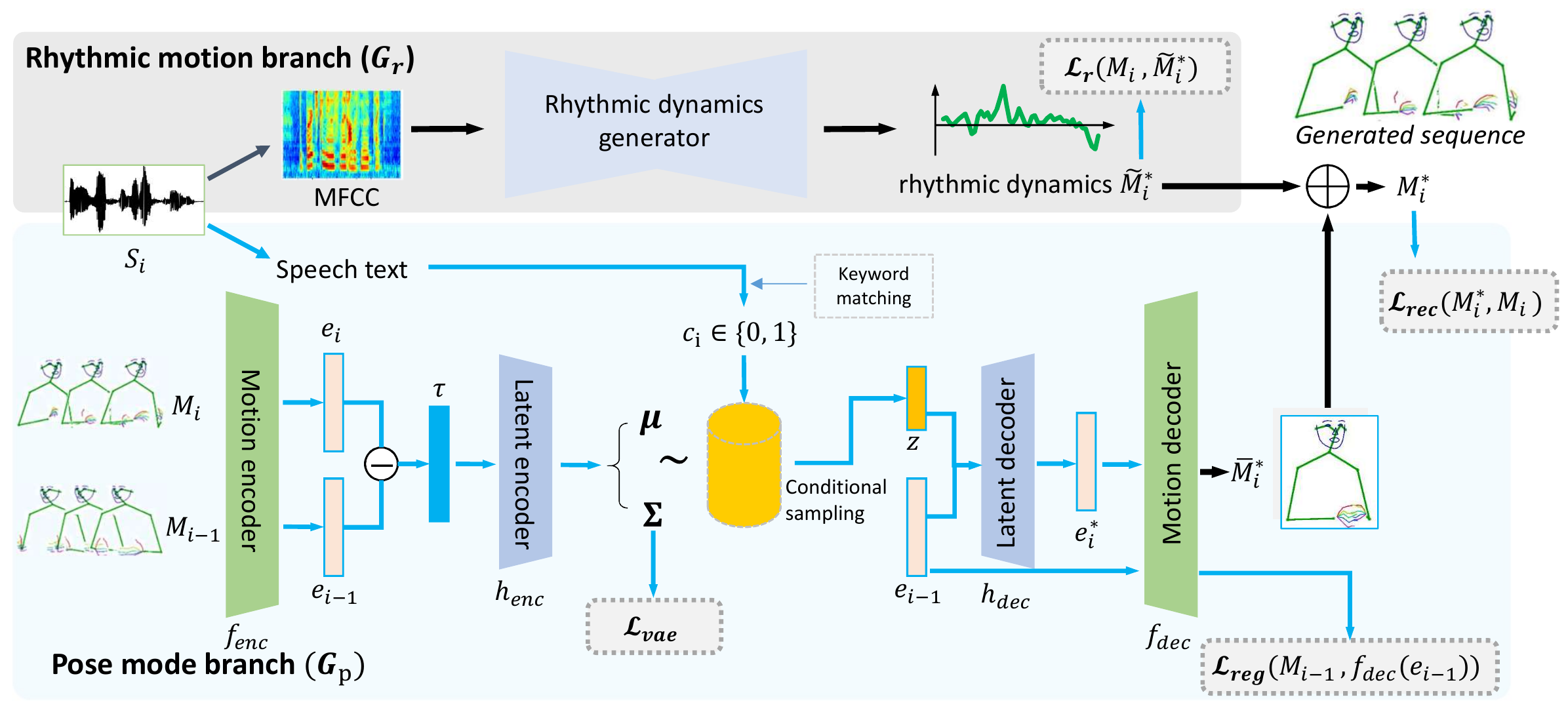}
  \caption{The framework of our freeform co-speech motion generation model (FreeMo) during training. The whole framework consists of two branches, a rhythmic motion branch that learns the mapping from speech audio to rhythmic dynamics, and a pose mode branch which is constructed as a CVAE by conditional sampling based on semantic signals in speech text to generate free-form pose modes.}
  \label{fig: framework}
\end{figure*}

Co-speech body motion generation is to generate the corresponding motion sequence, according to a given speech audio. To this end, a mapping from speech to body motion is required. Unfortunately, such mapping is highly non-deterministic and multi-modal. We approach this problem by decomposing the body motion into two complementary parts: pose modes and rhythmic dynamics. Correspondingly, our framework consists of two branches: a pose mode branch and a rhythmic motion branch, as shown in Fig.~\ref{fig: framework}.

% \subsection{Problem Formulation}
\subsection{Framework Overview}
Given a piece of speech audio $S$ and the corresponding body motion $M$, co-speech motion generation is to establish the mapping: $S\rightarrow M$. Without loss of generality, in this work, $S$ and $M$ are represented as MFCC \cite{MFCC} feature and human pose landmarks \cite{fang2017rmpe, li2018crowdpose, xiu2018poseflow}, respectively. To ease discussion, $S$ and $M$ are chunked into clips with small duration, i.e., $S = [S_1, S_2, ..., S_n]$, and $M=[M_1, M_2, ..., M_n]$, where each $S_i \in \mathbb{R}^{T\times D_S}$ or $M_i \in \mathbb{R}^{T\times D_M}$ corresponds to $T$ consecutive frames in the talk videos. $D_S$ and $D_M$ indicate the feature dimension of audio and motion respectively. 
With the Markov assumption, the problem turns to learn the mapping from $(S_i, M_{i-1})$ to $M_i$.

% At step $t$, the input is the audio of the current step $a_t$ and the generated motion sequence of the previous step $S_{t-1}^*$. The output is a keypoint sequence $S_t^* \in R^{T\times D}$, where $T$ is the length of the sequence and $D$ is the keypoint dimension. 

Learning such mapping is non-trivial due to the non-deterministic nature. Motivated by studies in linguistics and behavior \cite{gesture_and_speech_in_interaction_an_overview, gesture-visible_action_as_utterance, hand_movements, gesture_and_thought, Hand_and_Mind}, we decompose the body motions $M_i$ into two parts: 
\begin{equation}
    M_i=\overline{M}_i + \widetilde{M}_i,
    \label{Eq: additive}
\end{equation}
where $\overline{M}_i$ encodes the primary pose mode, and $\widetilde{M}_t$ embeds the rhythmic dynamics of $M_i$. 
Accordingly, our framework is composed with two major branches, where the pose mode branch ($G_p$) controls the primary postures ($\overline{M}_i$), conditioning on the audio $S_i$ and last step motion $M_{i-1}$:
\begin{equation}
    \overline{M}_i^* = G_{p}(M_{i-1}, S_i),
    \label{Eq: pose mode gene}
\end{equation}
and the rhythmic motion branch ($G_r$) is responsible for the rhythmic dynamics ($\widetilde{M}_t$) synced with the speech prosody:
\begin{equation}
    \widetilde{M}_i^* = G_{r}(S_i).
\end{equation}
We put a superscript (*) to denote the generated results. With such decomposition, the pose mode branch is only responsible for the pose mode generation, and the rhythmic motion branch learns the rhythmic dynamics independent to the primary pose.

% Our framework takes data samples as $\{a_i, S_{i-1}, S_i\}$, by randomly sampling from training videos, and the final generation result for $a$ is the temporal stack of the results at each time stamp. 

\subsection{Pose Mode Generation}

%At time step $i$, the pose mode branch generates a sequence $\overline{S}_i^*$ that embeds the primary pose mode of $S_i$. 
During a talk speech, the speakers' primary postures are mostly habitual and have no conventionalized linguistic meaning. We do not apply specific constraints on the forms of pose modes, but introduce a conditioning sampling trick to switch among learned pose modes. 

We first introduce a label $c_i \in \{0, 1\}$ to indicate whether there is a pose mode change from $M_{i-1}$ to $M_i$, and then learn a latent space to generate transitions among pose modes.
%, which is divided into sub-spaces according to $c_i$.
%During inference, we infer $c_i$ based on the speech text in $a_i$, and conditionally sample from the learned sub-space. 
Conditioning on $c_i$, the generated $\overline{M}_i^*$ either keeps the pose of last step, or transits into a new one.
Specifically, a conditional variational autoencoder (CVAE) is introduced, whose latent space encodes the modes transition between training pairs $(M_{i-1}, M_i)$ given $c_i$. 
% During inference, when $c_i=0$, we sample the latent code from training samples where $S_{i-1}$ and $S_i$ are in a same pose mode; when $c_i=1$, we sample the latent code from training samples where $S_{i-1}$ and $S_i$ are in different pose modes. 
%To do this, for each training sample, we first attach a pseudo label $d_t \ \in \{0, 1\}$ which indicates whether $S_{t-1}$ and $S_t$ are in a same pose mode or not. $d_t$ is automatically obtained by calculating the relative change of finger positions in $S_{t-1}$ and $S_t$, as shown in Fig.~\ref{fig: finger}.

%\noindent\textbf{Pose mode}. 
Following the convention \cite{MT-VAE}, the transition feature can be represented as the difference of encoded motions:
\begin{equation}
    \tau = e_i - e_{i-1},
\end{equation}
where $e_i = f_{enc}(M_i)$, $e_{i-1} = f_{enc}(M_{i-1})$, and $f$ is the motion encoder network. The VAE encoder ($h_{enc}$) takes $\tau$ as the input and calculates the posterior distribution as $\mathcal{N}(\mu_\theta(\tau), \Sigma_\theta (\tau))$. The VAE decoder takes the sampled vector $z \sim \mathcal{N}(\mu_\theta(\tau), \Sigma_\theta (\tau))$ and $e_{i-1}$ to recover $e^*_{i}$, which is then decoded as $\overline{M}^*_i=f_{dec}(e^*_i)$. 

% \begin{figure}
%   \centering
%   \includegraphics[width=0.89\linewidth]{figures/auto_label.pdf}
%   \caption{Each training sample $S_{t-1}, S_t$ is attached to a binary label $d_t\in\{0,1\}$ which indicates $S_{t-1}$ and $S_t$ are in a same pose mode or not. $d_t$ is obtained by calculating the relative change of finger positions.}
%   \label{fig: finger}
% \end{figure}

\noindent\textbf{VAE Latent Loss}. A VAE loss is introduced for conditional sampling. For each training sample $\{M_{i-1}, M_i, S_i\}$, $c_i$ indicates whether there is a pose mode switch at time $i$. Ideally, $c_i$ is closely related to the semantics (e.g., topic shifts) of speech. Practically, this can also be pseudoly labeled by calculating the extent of hand position change between $M_{i-1}$ and $M_i$.
When $c_i = 1$ (with pose change), we minimize the KL divergence between estimated posterior distribution $\mathcal{N}(\mu_\theta(\tau), \sigma_\theta (\tau))$ and the prior $\mathcal{N}(\mathbf{0}, \mathbf{I})$ \cite{VAE}, as $M_{i-1}$ and $M_i$ are with different modes:
\begin{equation}
    \mathcal{L}_{1} = KL(\mathcal{N}(\mu_\theta(\tau), \sigma_\theta (\tau)) || \mathcal{N}(\mathbf{0}, \mathbf{I})).
\end{equation}
When $c_i=0$ (without pose change), we use a zero vector as the latent transition representation $z$ for all training samples (with $c_i=0$), that is, $e^*_i=h_{dec}(\textbf{0}, e_{i-1})$. To keep the consistency of the latent encoder and latent decoder, we introducing the following constraint: 
\begin{equation}
    \mathcal{L}_{0} = \left\| \mu_\theta(\tau) \right\| + \left\| \Sigma_\theta(\tau) \right\|.
\end{equation}
Note that while $M_i$ and $M_{i-1}$ are in a same pose modes, they may have different rhythmic dynamics. Mapping all such samples into a same latent representation makes the encoder ignore the differences in rhythmic dynamics but focus on pose modes. 
The complete VAE latent loss can be written as:
\begin{equation}
    \mathcal{L}_{vae} = \mathbb{I}(c_i=0)\mathcal{L}_{0} + \mathbb{I}(c_i=1)\mathcal{L}_{1}.
\end{equation}

\noindent\textbf{Regularizing the embedding space}. Our pose mode branch is formulated as a conditional motion prediction. However, only deterministic data samples are available for training (for each $M_{i-1}$, the subsequent $M_i$ is fixed). As a result, the latent code $z$ can be easily ignored, and the model degenerates to deterministic prediction, i.e., $\overline{M}_i^*=f_{dec}(h_{dec}(z, e_{i-1})) \rightarrow \overline{M}_i^*=f_{dec}(e_{i-1})$, where $h_{dec}$ is the VAE decoder.
%With the assumption that $e_{i-1}$ is the embedding vector of $S_{i-1}$, the expected outputs of the decoder $g$ is $S_{i-1}=g(e_{i-1})$. 
We regularize the embedding space of the motion encoder $f$ and the motion decoder $g$ as:
\begin{equation}
    \mathcal{L}_{reg} = \left \| M_i-f_{dec}(e_i) \right \| + \left \| M_{i-1}-f_{dec}(e_{i-1}) \right \|.
\end{equation}
%Here $f$ and decoder $g$ are expected as an auto-encoder, instead of motion predictor. 
This prevents the motion decoder $g$ from ignoring the latent code $z$ and basing its predictions on $e_{i-1}$.

\subsection{Rhythmic Motion Generation}
Besides the primary posture, the temporal dynamics of the body motion is also an essential part in speech-motion correlation \cite{Hand_and_Mind}. The rhythmic dynamics in co-speech motions are simple and fast movements driven by the prosodic events in speech, contributing to the perceived prominence of temporally aligned speech and the sense of audiovisual prosody. We introduce the rhythmic motion branch to further match the dynamics with the speech prosody. 
% It takes the audio feature of current step as input, and output a sequence $\widetilde{S}^*_t$ that embeds the rhythmic dynamics of $S^*_t$. 
% As can be observed in Fig.~\ref{fig: finger}, the prosody-driven rhythmic dynamics mainly appear when $d_t=0$. When $d_t=1$, the dynamics of the motion sequences are mainly introduced by the pose mode transition process. Therefore, the rhythmic motion branch is only activated when $c_t=0$ (inference) or $d_t=0$ (training), otherwise it outputs a zero sequence.
%
In this work, we use a convolutional network as the rhythmic dynamics generator. 

\begin{figure}
  \centering
  \includegraphics[width=\linewidth]{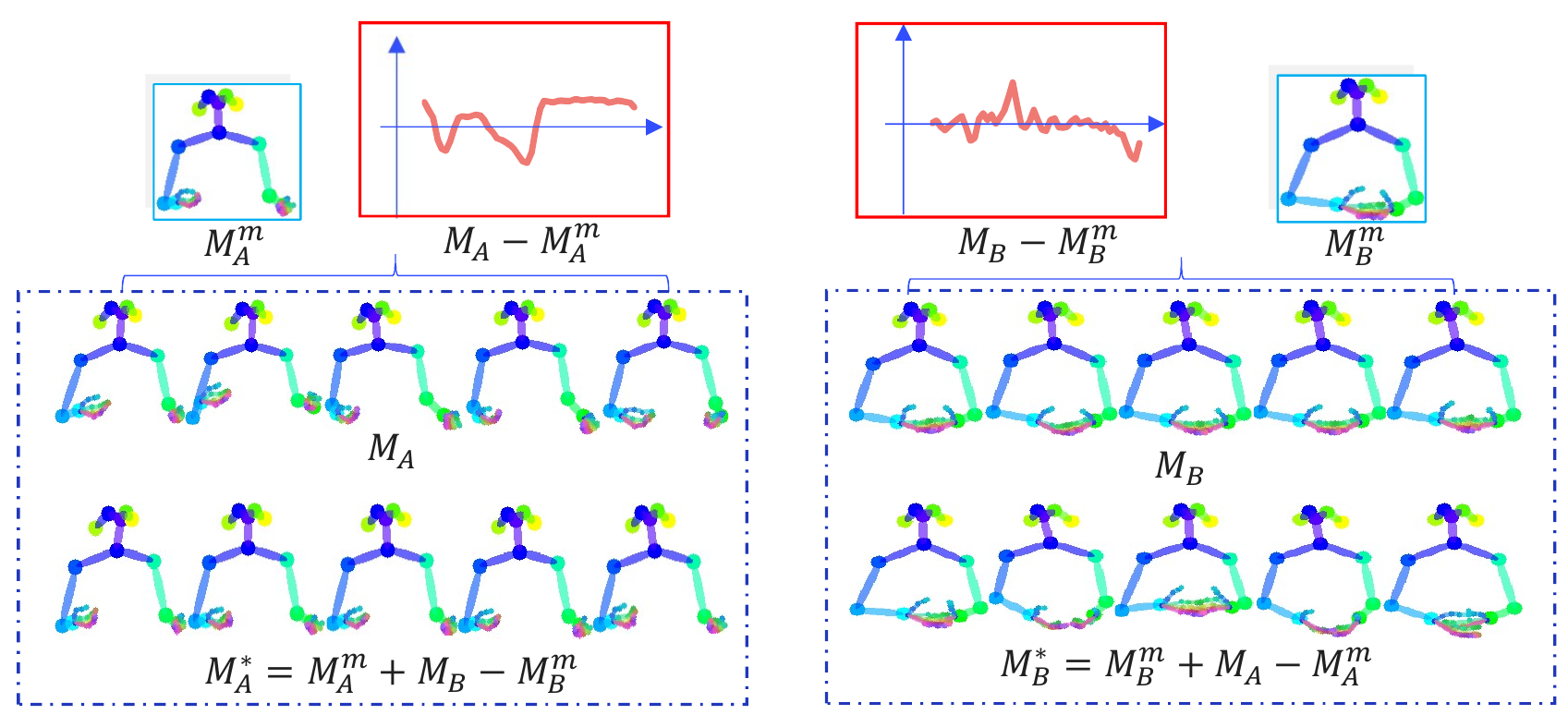}
  \caption{Disentangling rhythmic dynamics and primary posture in body motions. $M_A$ and $M_B$ are two random motion sequences sampled from our training set. The red boxes denote the offset to the mean posture of the motion sequence. We can control the motion dynamics without affecting the primary postures by exchanging the offset in the motion sequences. See supplement files for better temporal illustration.}
  \label{fig: visualiation_real}
\end{figure}

\noindent\textbf{Rhythmic motion loss}.
%For a same input $a_i$, the generated pose mode $\overline{S}_i$ by the pose mode branch can be different (speakers can speak the same sentence in different postures). 
The rhythmic motion branch learns the rhythmic dynamics ($\widetilde{M}^*_i$) independent to the primary poses. The corresponding loss function is defined as:
\begin{equation}
    \mathcal{L}_{r}=\left \| \widetilde{M}^*_i-(M_i - M_i^{m}) \right \|,
    \label{Eq: rhythm}
\end{equation}
where $M_i^m$ is the arithmetic mean of motions in $M_i$ (over $T$ frames). $\mathcal{L}_{r}$ ensures the generated result $\widetilde{M}_i^*$ with the proper offset to the mean posture, which helps control the motion dynamics without affecting the primary posture.
To verify this effect, we conduct a test to swap the primary posture and rhythmic dynamics as shown in Fig.~\ref{fig: visualiation_real}. Given two motion sequences $M_A$ and $M_B$, we exchange their motion dynamics as $M_A^* = M_A^m + (M_B - M_B^m)$ and $M_B^* = M_B^m + (M_A - M_A^m)$. As shown, while the motion dynamics are exchanged, the principle postures in $M_A$ and $M_B$ are preserved in $M_A^*$ and $M_B^*$, and the motion fidelity is not affected. % Therefore, we rely on $\mathcal{L}_{r}$ to enable posture-agnostic dynamics controlling.

\subsection{Overall Loss Function}

We apply the reconstruction loss on the final result $M^*_i = \overline{M}_i^* + \widetilde{M}_i^*$:
\begin{equation}
    \mathcal{L}_{rec} = \left \| M^*_i - M_i \right \|.
\end{equation}
%
% In addition, as we recurrently generate motion sequence at each time step, it is important to ensure the temporal continuity at adjacent steps. We use a velocity loss \cite{MT-VAE} to ensure a smooth velocity in $M_i^*$:
% \begin{equation}
%     \mathcal{L}_{vel}=\left \| v^*_i-v_i \right \|,
% \end{equation}
% where $v_i$ and  $v_i^*$ are the velocity of the ground-truth $M_i$ and generated sequence $M_i^*$, respectively.
%
Overall, the loss function is summarized as:
\begin{equation}
\begin{split}
    \mathcal{L}= \lambda_{1} \mathcal{L}_{rec} + \lambda_{2} \mathcal{L}_{vae}
     + \lambda_{3}\mathcal{L}_{r} + \lambda_{4} \mathcal{L}_{reg},
\end{split}\label{eq: total loss}
\end{equation}
where $\lambda_{1} \sim \lambda_{4}$ are the balancing weights for each loss.

\subsection{Inference Pipeline}

During inference, the pose mode branch implements the following function:
 \begin{equation}
     \overline{M}_i^*=G_p(M_{i-1}^*, S_i)=f_{dec}(h_{dec}(z, e_{i-1})),
 \end{equation}
where $z$ is a randomly sampled vector conditioned on $c_i$, and $c_i$ is inferred from the speech text in $S_i$ (e.g., with keyword matching). 
Specifically, when $c_i=0$, $z$ is a zero vector encoding the mode transitions where $M_{i-1}$ and $M_i$ are in a same pose mode; when $c_i=1$, $z$ is randomly sampled from $\mathcal{N}(\mathbf{0}, \mathbf{1})$, encoding the mode transition of training samples where $M_{i-1}$ and $M_i$ are in different pose modes. The final generation result for the whole input speech $S$ is the temporal stack of the generated motions at each step $[M^*_1,, M^*_2, ..., M^*_n]$, where $M_i^* = \overline{M}^*_i+\widetilde{M}^*_i$. 

\iffalse
\begin{algorithm}
% \DontPrintSemicolon
  
  \KwInput{Speech audio $a = [a_1, a_2, ..., a_n]$; initial pose $S^*_0$}
  \KwOutput{Generated motion sequence $S^*=[S^*_1, S^*_2, ..., S^*_n]$}
  \For{$i$ from $1$ to $n$}{
    \tcc{Infer $c_i$ from $a_i$ using key-word matching}
    $c_i=map(a_i)$ \\
    \tcc{Generate pose mode}
    $e_{i-1}^*=f(S^*_{i-1})$ \\
    \If{$c_i=0$}{
    $z=\mathbf{0}$}
    \ElseIf{$c_i=0$}{
    $z\sim\mathcal{N}(\mathbf{0}, \mathbf{1})$}
    $e^*_i=h_{z\rightarrow e}(e_{i-1}^*, z)$ \\
    $\overline{M}^*=g(e^*_i)$ \\
    \tcc{Generate rhythmic dynamics}
    $\widetilde{M}^*_i=\mathcal{G}_r(a_i)$ \\
    \tcc{Generated sequence at step $i$}
    $S_i^*=\overline{S}^* + \widetilde{M}^*$
  }

\caption{Inference code}
\label{algo: inference}
\end{algorithm}
\fi

\section{Experimental Results}

%We first introduce our constructed co-speech motion dataset \textit{TEDGesture}, and then empirically study our proposed method for freeform co-speech motion generation. 

% We evaluate our FreeMo with quanlitative, quantitative and user study. 

\subsection{Experimental Setup}
\begin{figure}[t]
  \centering
  \includegraphics[width=\linewidth]{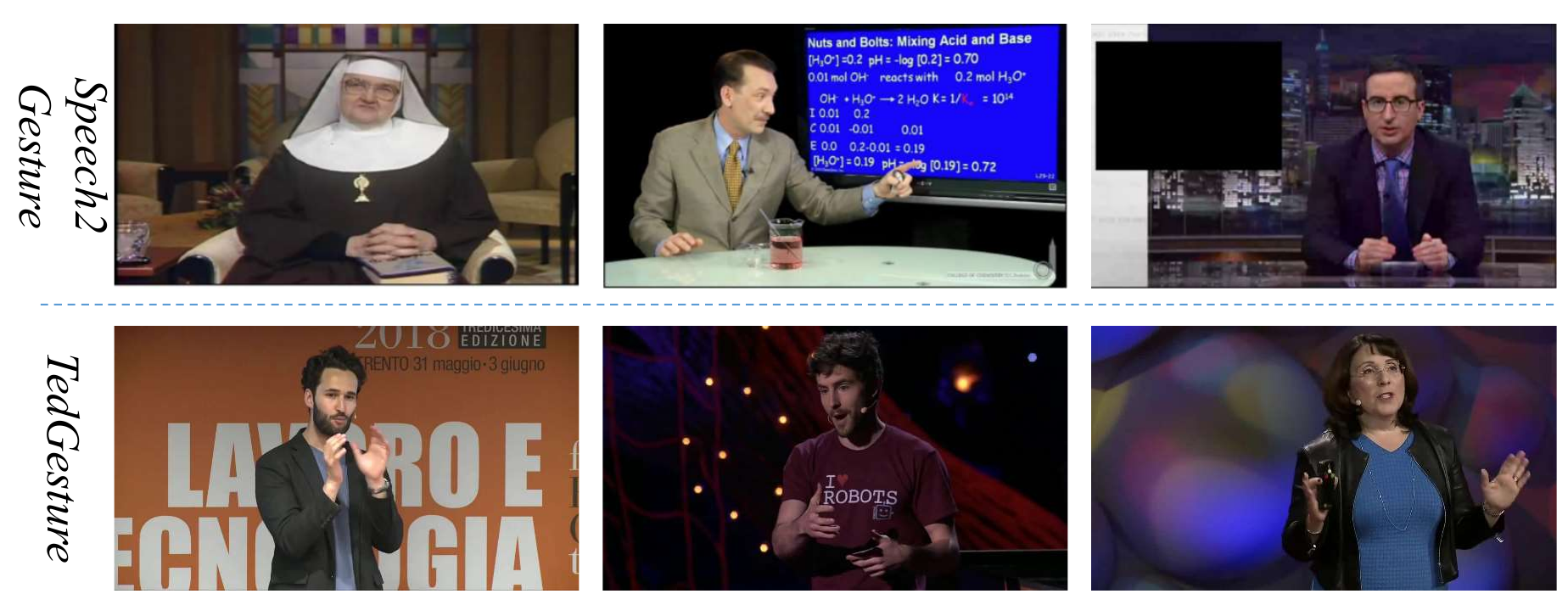}
  \caption{Example videos in \textit{Speech2Gesture} dataset (top) and \textit{TEDGesture} dataset (bottom).}
  \label{fig:snapshot}
\end{figure}

\noindent\textbf{Datasets.}
% Co-speech motions are known to be idiosyncratic \cite{Hand_and_Mind}. Different speakers have different styles for making body motions during speech. 
Following the convention \cite{speech2gesture, speech_drives_templates, style_transfer}, we also test on the the \textit{Speech2Gesture} dataset \cite{speech2gesture}, which contains speaker-specific videos of Television anchors. However, most videos are TV shows with heavy interference from the environment (e.g., sound from audience, front desk), and the body motions are constrained as the speakers often seat in chairs or lean against desks (Fig.~\ref{fig:snapshot}). % Videos in the form of keynote speeches usually contains whole body motions, and the motion in high-quality keynote speeches are also more contagious than the staid body language in the Speech2Gesture dataset. 

Therefore, we include another \textit{Ted Gesture} dataset \cite{robots_learn_social_skills} to evaluate on videos of keynote talks. Following the setting as in \cite{speech2gesture}, we add more videos collected from YouTube to ensure enough data for each speaker. %As a result, 1,979 video segments from 71 videos of 20 speakers are adopted for evaluation. 
We split these segments as 80\% for training, 10\% for testing, 10\% for validation.
Fig.~\ref{fig:snapshot} shows some snapshots of Speech2Gesture and TEDGesture.

% Besides the Speaker-specific co-speech motion (\textit{Speech2Gesture}) dataset \cite{speech2gesture}, we include another TEDGesture dataset for experiments. As shown in Fig.~\ref{fig:snapshot}, speakers in Speech2Gesture are often seated in chairs, leaning against desks, such that their motions are heavily constrained (e.g., only twisting the wrists). Moreover, most of the videos in Speech2Gesture are television shows with heavy interference from the environment (e.g., the sound of audience, interactions between speaker and the desk). 
%
% Therefore, we additional collect a talk-oriented dataset \textit{TEDGesture} for co-speech motion evaluation. Videos in the form of keynote speeches usually contains whole body motions, and the motion in high-quality keynote speeches are also more contagious than the staid body language in the \textit{Speech2Gesture} dataset. We first pick several experienced speakers from TED.com, and then collect videos from YouTube to ensure there is enough data for each speaker. 
% % All the videos in the collected dataset are stand-up talks in the form of keynote speeches. 
% In total, we get 1,979 segments from 71 videos of 20 speakers. We split these segments as 80\% for training, 10\% for testing, 10\% for validation. Fig.~\ref{fig:snapshot} shows the snapshots of \textit{Speech2Gesture} and our constructed dataset \textit{TEDGesture}.

\noindent\textbf{Compared Methods.}
We compare with the following state-of-the-art methods:
\begin{itemize}
\item Audio to Body Dynamics (Audio2Body) \cite{audio_to_body_dynamics} adopts an RNN network for audio to motion translation.
\item Speech2Gesture (S2G) \cite{speech2gesture} proposes a convolutional network for speech to gesture generation. %Similar to A2B, it also learns a direct mapping from audio to motion sequence.
\item Speech Drives Template (Tmpt) \cite{speech_drives_templates} learns a set of gesture templates to relieve the ambiguity of the mapping from speech to body motion. 
\item Trimodal-Context (TriCon) \cite{trimodal} adopts an RNN network that takes (audio, text, speaker) triplet as input. Speaker ID specifies the styles for generated motions.
% \item MoGlow (MG) \cite{style_control} adapts a probabilistic synthesis method MoGlow \cite{moglow} for co-speech motion generation, which mainly improves the motion diversity. 
\item Mix-StAGE \cite{style_transfer} learns a unique style embedding for each speaker using a mixture of generation models.
\end{itemize}

\noindent\textbf{Evaluation Protocol.}
Co-speech motion generation is essentially a non-deterministic prediction task, as multiple motion sequences are plausible for a same speech audio. Metrics that only evaluate against one ground-truth motion (e.g., \textit{L1} loss \cite{speech2gesture}) are somewhat biased. In this work, three quantitative metrics \cite{AI_choreographer} are considered for evaluating the generated motions: 1) \textit{syncing} between speech and motion, 2) \textit{diversity} of motions, and 3) \textit{quality} of motions as compared to real ones. Besides, qualitative comparison and user study are included as subjective evaluations. 

\subsection{Quantitative Results}
% This section presents the quantitative evaluation results.
\noindent\textbf{Syncing.}
The synchronization between generated motions and speech is one of the most essential factors indicating how well the model captures the relation of speech and motion.
Landmark velocity difference (LVD) \cite{makeittalk} is adopted to evaluate the syncing with speech, as we do not expect to generate motions exactly the same as the ground-truth. % as multiple motions can go well with the speech. LVD encourages such motions as long as they share a similar dynamics with the ground-truth. 

\begin{table*}[t]
\renewcommand{\arraystretch}{1.05}
\centering
%\resizebox{\textwidth}{!}{
\small
\begin{tabular}{l|cccccccccc|c}
\hline
            & Amel  & Bill & Christina & Dan    & Dena & Enric & Kelly & Keller & Sara & Stanley & \textbf{Average}$\downarrow$ \\
\hline
\hline
Last-step Velocity & 15.9 & 34.0 & 26.3    & 21.9 & 20.8 & 13.5 & 17.6 & 21.3 & 14.9 & 17.9 & 20.4 \\
Mean Velocity  & 16.1 & 12.8 & 10.7    & 15.9 & 11.4 & 14.1 & 11.2 & 11.4 & 12.0 & 10.8 & 12.6 \\
\hline
Audio2Body \cite{audio_to_body_dynamics}             & 23.7 & 18.6 & 19.8    & 23.0 & 27.8  & 20.1 & 21.5 & 23.0 & 19.3 & 22.4 & 21.9 \\ 
TriCon \cite{trimodal}          & 16.6 & 16.5 & 20.2    & \textcolor{blue}{13.7} & 16.0  & 13.8 & 11.5 & 13.6 & 14.1 & 11.0 & 14.7 \\ 
S2G \cite{speech2gesture}              & 12.8 & \textcolor{red}{11.3} & 19.4 & 22.0  & 12.2 & 11.9 & \textcolor{blue}{10.6} & \textcolor{blue}{10.7} & \textcolor{blue}{11.8} & \textcolor{blue}{10.0} & 13.3 \\
Mix-StAGE \cite{style_transfer} & \textcolor{blue}{11.8} & 16.0 & \textcolor{blue}{9.9} & 16.5 & \textcolor{blue}{9.4} & \textcolor{blue}{10.5} & 17.3 & 11.4 & 12.2 & 14.7 & 13.0 \\
Tmpt \cite{speech_drives_templates} & 14.6 & 11.5 & 10.1 & 13.7 & 13.3 & 10.9 & 10.7 & 11.6 & 11.9 & 12.3 & \textcolor{blue}{12.1} \\
\hline
FreeMo (ours)             & \textcolor{red}{11.7} & \textcolor{blue}{11.4} & \textcolor{red}{9.2} & \textcolor{red}{12.7} & \textcolor{red}{8.0} & \textcolor{red}{9.7} & \textcolor{red}{10.4} & \textcolor{red}{10.3} & \textcolor{red}{11.0} & \textcolor{red}{9.0} & \textcolor{red}{10.3} \\
% Ours-pose mode      & 13.9 & 13.2 & 11.0    & 14.2 & 11.2 & 11.6 & 10.2 & 11.4 & 11.6 & 9.2 & 11.8\\
% \hline
% Ours-multi speaker    & 14.1 & 13.3 & 10.9    & 15.7 & 11.1  & 14.3 & 10.2 & 12.9 & 12.8 & 9.8 & 12.5\\
\hline
\end{tabular}
\caption{Quantitative comparison on \textit{TEDGesture} dataset for syncing between speech and motions using LVD (lower is better). The best and second results are marked in red and blue, respectively. 
%All models are speaker-specific, i.e., the results for each ID are obtained using the model trained on the speech videos of the corresponding speaker.
}\label{tab:comparisons}
%}
\end{table*}

\begin{figure*}[ht]
\begin{minipage}[t!]{.675\linewidth}
\small
\centering
\renewcommand{\tabcolsep}{3.25pt}
\renewcommand{\arraystretch}{1}
\resizebox{\linewidth}{!}{
\begin{tabular}{l|ccccccc|c}
\hline
            & Almaram  & Angelica & Chemistry & Ellen  & Oliver & Seth & Shelly & \textbf{Average}$\downarrow$ \\
\hline
\hline
Last-step Velocity                         & 17.5   & 16.6   & 11.1    & 18.5  & 17.4  & 19.2 & 33.1 & 19.1  \\
Mean Velocity                              & 10.3   & 7.4    & 6.1     & 9.0   & 7.2   & 5.1  & 16.1 & 8.7  \\
\hline
Audio2Body \cite{audio_to_body_dynamics}          & 14.5   & 23.1   & 8.3     & 10.0  & 14.1  & 11.5 & 17.1 & 14.1 \\
TriCon \cite{trimodal}                     & \textcolor{blue}{8.6}    & 9.3    & 5.2     & \textcolor{blue}{7.8}   & 5.7   & \textcolor{blue}{4.1}  & 13.8 & 7.8  \\ 
S2G \cite{speech2gesture}                  & 11.4   & \textcolor{red}{5.5}    & \textcolor{red}{4.4}     & 8.6   & \textcolor{red}{5.2}   & \textcolor{red}{3.9}  & \textcolor{red}{11.3} & \textcolor{blue}{7.2}  \\
Mix-StAGE \cite{style_transfer}            & 11.2   & 11.8   & \textcolor{blue}{4.7}     & 9.6   & 11.9  & 9.2  & 15.7 & 10.6 \\
Tmpt \cite{speech_drives_templates}& 9.6    & 6.2    & 4.9     & 7.9   & \textcolor{blue}{5.3}   & 4.2  & 13.1 & 7.3 \\
\hline
FreeMo (ours)                                       & \textcolor{red}{7.6}    & \textcolor{blue}{5.6}    & 4.8     & \textcolor{red}{6.8}   & 5.5   & 4.3  & \textcolor{blue}{12.4} & \textcolor{red}{6.7}  \\
\hline
\end{tabular}%}
}
\captionof{table}{Quantitative comparison on \textit{Speech2Gesture} dataset for syncing between speech and motions using LVD (lower is better). The best and second results are marked in red and blue, respectively. 
%All the models are speaker-specific.
}\label{tab:comparison_s2g}%, i.e., the results for each ID are obtained using the model trained on the speech videos of the corresponding speaker.}\label{tab:comparison_s2g}

\end{minipage}
~~~
\begin{minipage}[t!]{.325\linewidth}
  \centering
  \renewcommand{\arraystretch}{1.2}
\small
\resizebox{\linewidth}{!}{
\begin{tabular}{l|cc}
\hline
Method & Diversity$\uparrow$ & Quality$\uparrow$  \\
\hline
\hline
Audio2Body \cite{audio_to_body_dynamics} & \textit{NA} & 0.375 \\
S2G \cite{speech2gesture} & \textit{NA} & 0.163 \\
Mix-StAGE \cite{style_transfer} & \textit{NA} & 0.382 \\
Tmpt \cite{speech_drives_templates} & \textit{NA} & 0.178 \\
TriCon \cite{trimodal} & \textit{NA} & 0.278 \\
\hline
FreeMo (ours) & 0.160 & \textbf{0.502} \\
\hline
\end{tabular}
}
\captionof{table}{Quantitative evaluation on the quality and diversity. \textit{NA}: the method essentially learns a deterministic mapping and thus diversity is not applicable.}\label{tab:diversity}
\end{minipage}%
%\vskip -2.em
\end{figure*}

Tables~\ref{tab:comparisons} and \ref{tab:comparison_s2g} show the LVD scores of different models on TEDGesture and Speech2Gesture datasets, respectively. For completeness, we also include two direct data-driven baselines \cite{MT-VAE}: 1) \textit{Last-step Velocity} that uses the initial velocity of the ground-truth motion as the prediction; 2) \textit{Mean Velocity} that uses the average velocity of the ground-truth motion sequence as the prediction. Note that \textit{Mean Velocity} is actually a strong baseline, when comparing the LVD between ground-truth and generated motions.

As shown, FreeMo outperforms other baselines, suggesting better syncing ability in relating audio and motion. Our model shows strong performances on all speakers as evidenced by the low LVD scores, while other baselines fail to sync on some speakers. Our model is more robust to different speaker styles, leading to lower averaged LVD scores. Note that the absolute LVD scores on Speech2Gesture are smaller than those on TEDGesture, since body motions in Speech2Gesture are much simpler and static (Fig.~\ref{fig:snapshot}).

\noindent\textbf{Diversity and Quality.}
We further evaluate our method in terms of diversity and quality. Specifically, we measure the diversity by calculating the average distances among 64 motion sequences generated from a same initial posture for each test audio.
To measure the quality (fidelity) of generated motions, we train a binary classifier to discriminate real samples from fake ones and use the prediction score on the test set as the quality of the generated motions  \cite{mix-match}.

Table~\ref{tab:diversity} shows the comparison results of diversity and quality on different methods. S2G, Mix-StAGE, Tmpt, Audio2Body and Tricon are all essentially based on deterministic models. Therefore the diversity is not applicable, i.e., fixed output for the same input. Our FreeMo reaches a diversity of 0.163 due to the conditional sampling technique. As a comparison, if we manually set different style inputs for TriCon to generate diverse motions, the diversity score reaches 0.193, which is slightly higher than our FreeMo. 

The quality of our model is much better than all baselines, mimicking moves similar to real speakers. This demonstrates the ability to generate both diverse and plausible co-speech motions, which can also be observed in our user study. Note that better syncing leads to degenerated diversity, and these scores are better jointly considered. 
% Also note that the diversity of FreeMo and TriCon are slightly different. While both of them rely on randomly sampled vectors for generation, TriCon uses the sampled vectors to specify the styles of motion, and we use the sampled vectors to generate the transition among pose modes.

\begin{figure}[t]
  \centering
  \includegraphics[width=\linewidth]{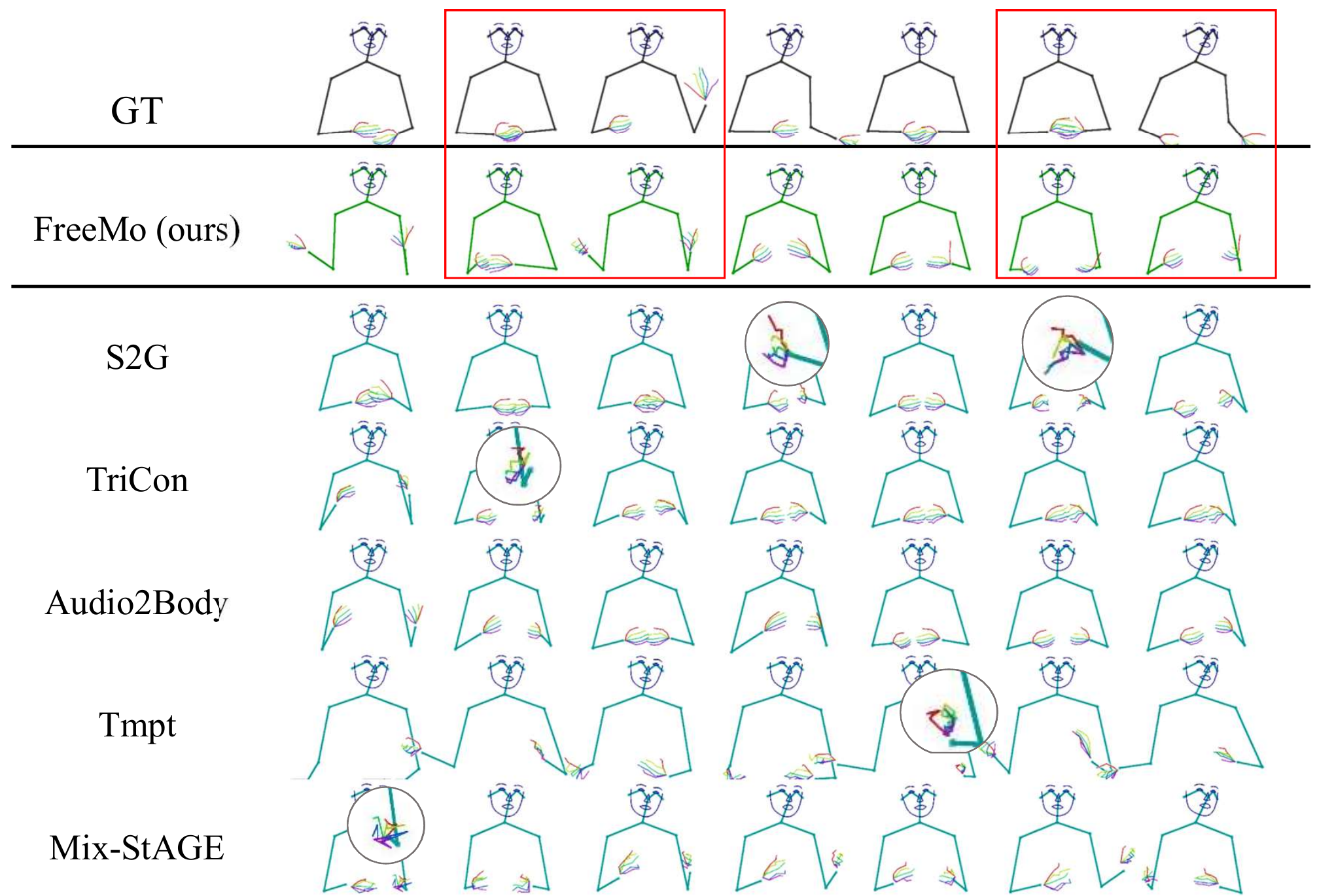}
  \caption{Qualitative comparisons on co-speech motion generation results. We highlight the unrealistic hand deformations appeared in the baseline methods. We also outline the pose changes in red boxes to show the expressiveness of our method. See supplementary material for more video results.}
  \label{fig:quality}
\end{figure}

\subsection{Qualitative Results}

Fig.~\ref{fig:quality} shows the qualitative comparisons across different models. Motions generated by the baseline methods contain hand deformations, while such deformations can hardly be found in our model. Intuitively, motions generated by S2G and TriCon typically are small movements around a certain posture. They lack clear pose transitions as appeared in the ground-truth (outlined in the red boxes), leading to less realistic and expressive motions. Such movements are difficult to synthesize through the direct audio to motion mapping.
Motions by Audio2Body show noisy dynamics and poor temporal continuity. Compared to these methods, our FreeMo generates more natural and expressive motions.

A common problem for co-speech motion generation is the severe jittering, which can be observed in almost all baselines (see supplementary file for temporal consistency). They directly translate speech to motion sequence, and thus the audio-conditioned dynamics and the primary postures are coupled and simultaneously inferred from the audio. Benefiting from the decomposition, our model has much stable generation quality. 

\begin{figure}[t]
  \centering
  \includegraphics[width=\linewidth]{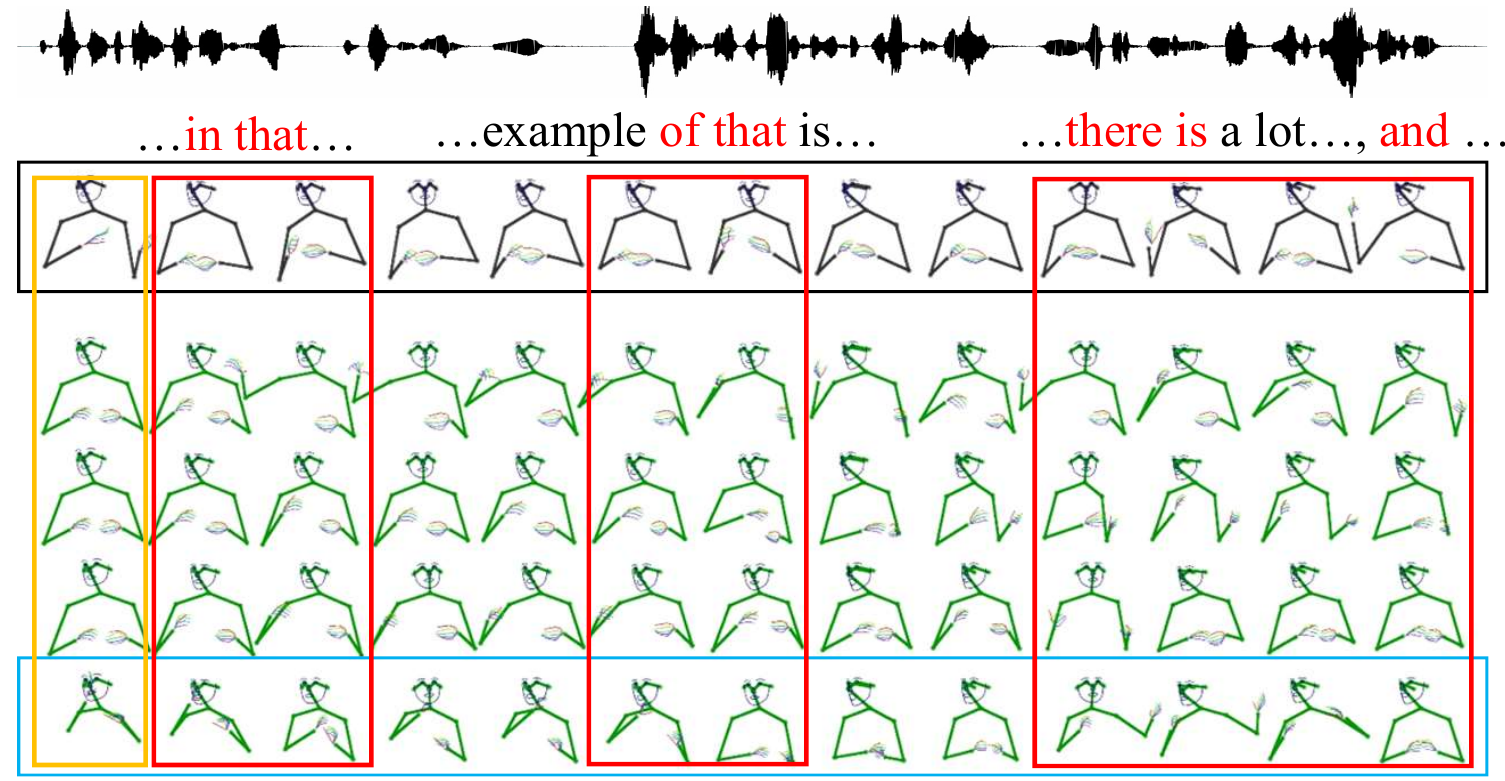}
  \caption{Multiple motion sequences for a same speech generated by our model. The red boxes highlight the occurrences of pose mode transitions in the generated motions and the ground-truth motions (first row). See supplementary material for video results.}
  \label{fig:diversity}
\end{figure}

To demonstrate the high diversity of our model, we provide multiple generation results of our model for the same speech in Fig.~\ref{fig:diversity}, including motions generated from a same initial posture and different initial postures. Our model generates diverse motion sequences from an arbitrary initial posture. It is worth noting that we do not rely on different initialization to obtain diverse results. This is achieved by the stochastic mode transitions of the pose mode branch. In the red boxes we highlight the transitions of pose modes. At each transition, the generated posture has a large degree of freedom, while the occurrences of transitions are synchronized with the ground-truth. %This makes the generated motions diverse while also well synced with the speech. 
Besides, the rhythmic branch plays a critical role in syncing the generated motion with the speech, such that the dynamics of generated sequence match well to the audio. %while these sequences may be in different postures. 

\subsection{Subjective Evaluation}

\begin{table}
\renewcommand{\arraystretch}{1.1}
\centering
\small
%\resizebox{0.8\linewidth}{!}{
\begin{tabular}{l|c|c}
\hline
 & TEDGesture & Speech2Gesture \\
\hline
\hline
Audio2Body \cite{audio_to_body_dynamics} & 2.09 & 2.11 \\
TriCon \cite{trimodal} &  2.36 & 2.89 \\ 
S2G \cite{speech2gesture} & 2.81 & 3.28 \\
Tmpt \cite{speech_drives_templates} & 4.09 & 4.55 \\
Mix-StAGE \cite{style_transfer} & 3.91 & 3.11 \\
\hline
FreeMo & \textbf{5.73} & \textbf{5.05} \\
\hline
\end{tabular}%}
\caption{Subjective user study on two datasets. The participants are asked to rank the videos generated by different models according to their preferences from highest (6) to lowest (1). Each row shows the averaged user preference rank for each method.}
\label{table: user study}
\end{table}

We further conduct subjective user study to evaluate the user preferences against several baselines. 
For each dataset, we randomly select 50 test audio clips. These audio clips are of different lengths from 10s to 30s. The participants are asked to rank the videos generated by different models according to their preferences from highest (6) to lowest (1). Ten participants attended in this user study, each of them watches the videos for 10 randomly chosen audio clips. Table~\ref{table: user study} reports the average preference scores. We can see that on both datasets, our proposed FreeMo has highest scores, preferred by most users. According to the participants' feedback, their decisions are mostly influenced by three aspects: 1) visual fidelity, e.g., hand deformations, motion continuity and speed; 2) the syncing to special prosodic events, such as emphasis, turning point in the speech; 3) motion expressiveness, e.g., motions with large and concise movements are preferred, while motions with small (or static) movements are less favored.

\begin{table}[t]
\renewcommand{\arraystretch}{1.1}
\renewcommand{\tabcolsep}{4pt}
\centering
%\resizebox{\linewidth}{!}{
\small
\begin{tabular}{cccccc}
\hline
$\mathcal{L}_r$  &  $\mathcal{L}_{reg}$ & multi-speaker & Syncing$\downarrow$  & Diversity$\uparrow$ & Quality$\uparrow$ \\
\hline
\hline
\xmark & \cmark  & \xmark  & 10.8 & 0.202 &  0.582  \\
\cmark & \xmark  & \xmark  & 10.6 & 0.125 &  0.583  \\
\cmark & \cmark  & \xmark & 10.3 & 0.160 & 0.502 \\
\cmark & \cmark  & \cmark  & 10.2 & 0.254 & 0.375\\
\hline
\end{tabular}
%}
\caption{Ablation study results on our FreeMo. \textit{multi-speaker} stands for training on multiple speakers simultaneously. These speakers are from a held-out set excluding speakers in the test set.}\label{tab: ablation}
%}
\end{table}

\subsection{Ablation Study}
We further run an ablation study on the losses ($\mathcal{L}_r$, $\mathcal{L}_{reg}$). Note that $\mathcal{L}_{rec}$ can not be removed for proper training, and removing $\mathcal{L}_{vae}$ degenerates the model to a deterministic mapping \cite{speech2gesture}.% (Tables~\ref{tab:comparisons}, \ref{tab:comparison_s2g}, \ref{tab:diversity}).

As shown in Table~\ref{tab: ablation}, removing $\mathcal{L}_r$ results in a worse syncing score. This is expected as the syncing of the posture transitions are preserved, while the dynamics of the generated sequences no longer match to the audio. The diversity and quality are higher than the full model, since the generated motions are less constrained and their rhythmic dynamics resemble the training samples. The generated motions for a same speech audio are also less similar to each other due to the worse syncing ability. Removing $\mathcal{L}_{reg}$ leads to a decline in diversity and syncing, as the model becomes more deterministic. These results also demonstrate that for co-speech motion generation, both semantics and prosody of the audio should be considered.

\subsection{Discussion on Cross-ID Generation}

% Previous methods generate speaker-specific co-speech motions, i.e., they train and test on a same speaker. Their performance are significantly worse when tested on a different speaker \cite{speech2gesture}. 

%Above methods are all based on speaker specific setting, i.e., one model for each speaker. 
For applications such as social robots and digital avatar, the driving audio usually are synthesized or recorded by voice actors. It is unlikely to train speaker-specific models for each speaker. Therefore, the generalization ability to different speakers is highly desired. Therefore we further analyze the performance on audio of unseen speakers.

\noindent\textbf{Same model on different audio.}
\begin{figure}
    \centering
    \includegraphics[width=\linewidth]{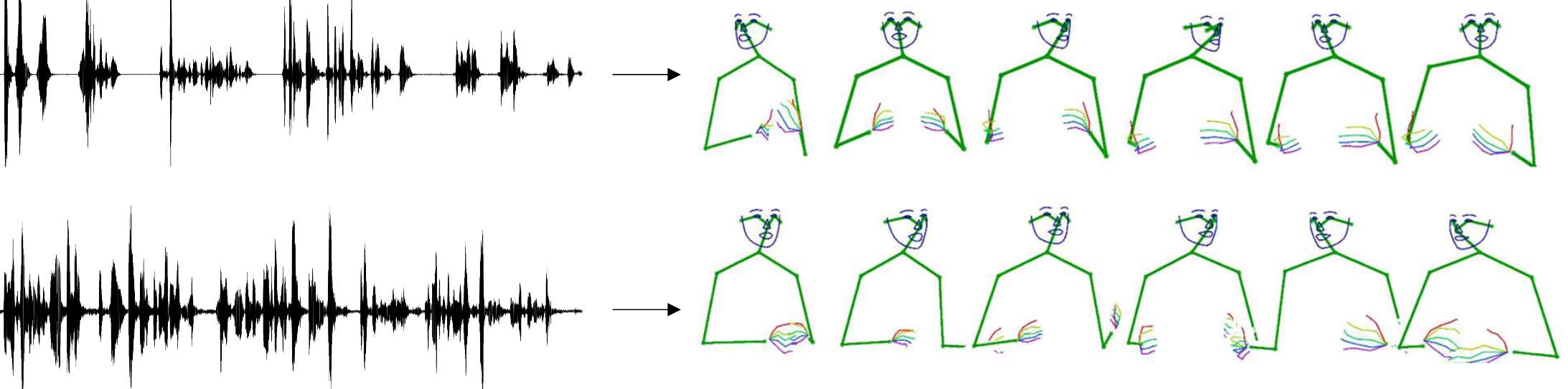}
    \caption{Same model on different audio. We select audios with different prosody, and compare the generation results of a same model on these audios.}
    \label{fig: same_model}
\end{figure}
The generalization ability to unseen speakers essentially reflects that the model generates body motions based on ID-independent speech features, such as semantics, intonation, etc., rather than overfit to ID-specific features such as timbre. To show this, we select audios from different speakers and compare the generation results by a fixed model. As shown in Fig.~\ref{fig: same_model}, for speech with low speed, the model generates soothing body movements. When the speed is faster, the movements will also be faster.

\begin{figure}
    \centering
    \includegraphics[width=\linewidth]{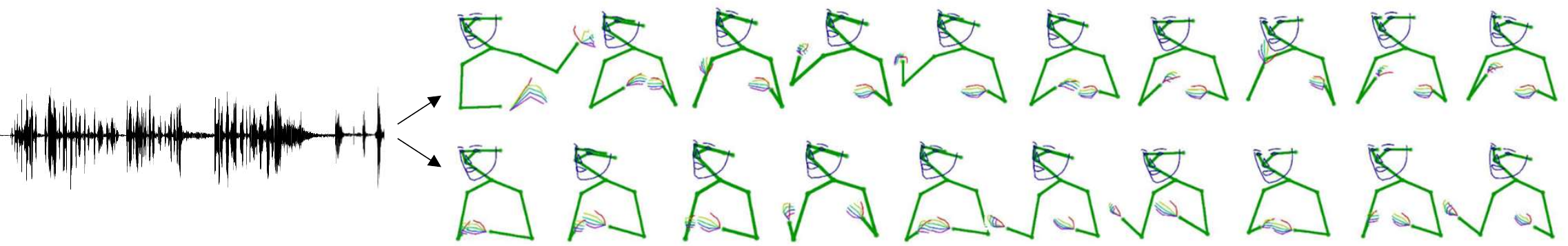}
    \caption{The generation results of different model on a same audio. While these models are trained on data of different speakers and they generate body motions with different postures, the generated motion dynamics are similar.}
    \label{fig:cross_id_2}
\end{figure}

\noindent\textbf{Different models on the same audio.}
Most existing works map speech to motion via conditioning on personal styles. Contrary to this, we do not attempt to fit the personal style of individual speakers. While different speakers may have their own styles, their motions contribute to the speech generally in a similar way \cite{gesture_and_speech_in_interaction_an_overview}.
The generated sequences for a same audio by different models are shown in Fig.~\ref{fig:cross_id_2}.
While these models are trained on different speakers and they generate body motions with different postures, the generated motion dynamics are roughly similar. This shows that our model captures the essential correlation between speech and motion that is shared across speakers. 

\begin{figure}[t]
  \centering
  \includegraphics[width=\linewidth]{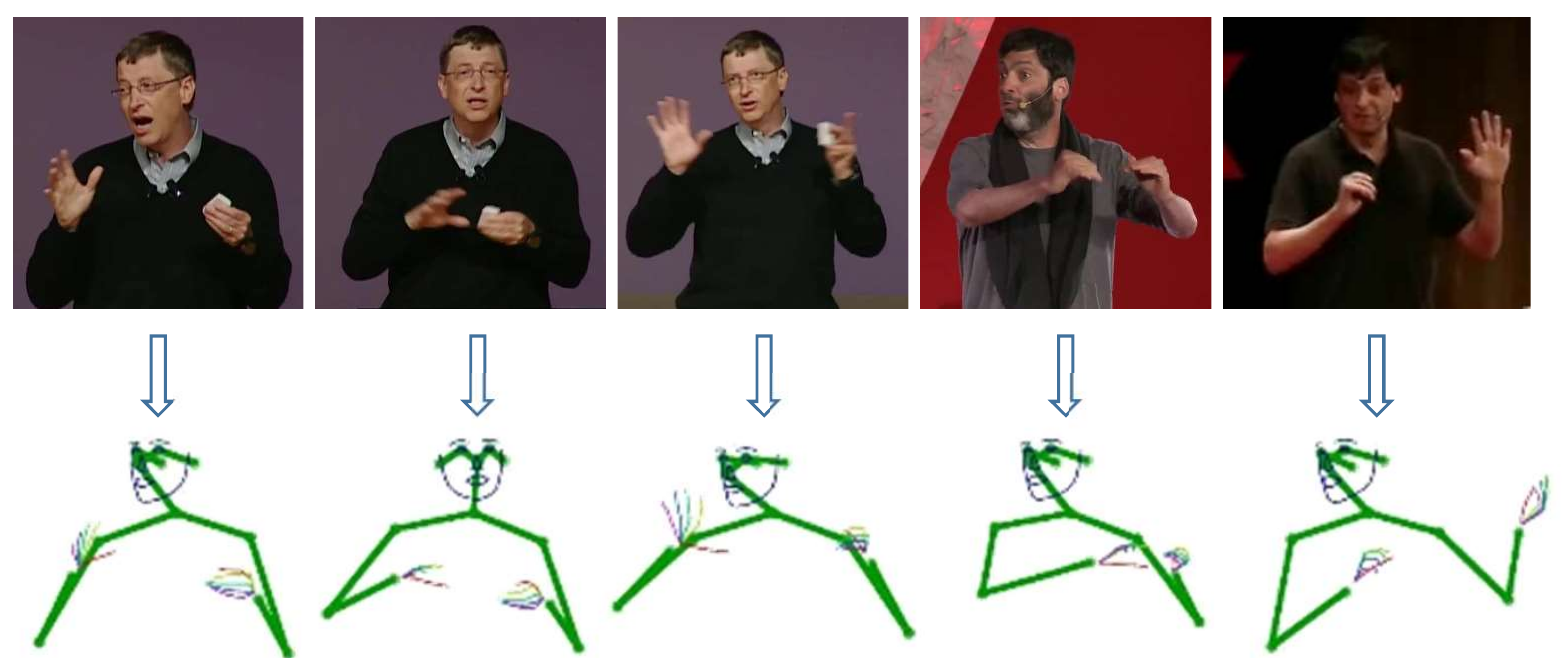}
  \caption{The original video frames of the nearest neighbors of the postures in the generated sequence trained on multi-speakers. The posture styles of different speakers are combined in a same generation result.}
  \label{fig: cross_id}
\end{figure}

\noindent\textbf{Training on multiple speakers.} 
The distribution of training data is critical to the generalization ability of deep learning models. However, the speech videos of a single speaker is very limited, due to personal habits. It is difficult to train a model with strong generalization ability using videos from a single speaker. Therefore, it is important to use data from different speakers to construct sufficient training data. To explore this, we use ten extra speakers videos as the training set. The results are shown in the last row of Table~\ref{tab: ablation}. The trained model has a comparable performance in syncing compared to the speaker-specific models (the third row in Table~\ref{tab: ablation}), and much better diversity. Our model successfully transits between the pose modes learnt from multiple speakers, showing highly diverse motions. 
Fig.~\ref{fig: cross_id} shows the original video frames of the nearest neighbors of postures in the generated sequence. There are multiple habitual postures of different speakers appearing in the same generation result. The posture style of different speakers are jointly modeled by our model, enabling higher generalization diversity. The lower quality score might be caused by the out-of-distribution problem, i.e., the extra ten speakers are not included while training the quality classifier. %generated motions are of posture styles different from the real samples for training the binary classifier.

% Based on the cross-id results, our model does not have to be speaker-specific and condition on personal styles.
% We attribute the generalization ability across speakers to the decomposition of pose modes and rhythmic motions.
% different speakers are similar in the way that the speech and motion correlates (i.e., their motions fulfil similar pragmatic functions), and are different in the habitual postures (i.e. pose modes).

%Based on these results, we show that co-speech motion generation models are not necessarily speaker-specific and have to be conditioned on personal styles. We attribute this generalization ability across different speakers to that different speakers are similar in the way the speech and motion are correlated (i.e., their motions fulfill similar pragmatic functions), and are different in the habitual postures.

\subsection{Video Rendering}

\begin{figure}[t]
   \centering
   \includegraphics[width=.99\linewidth]{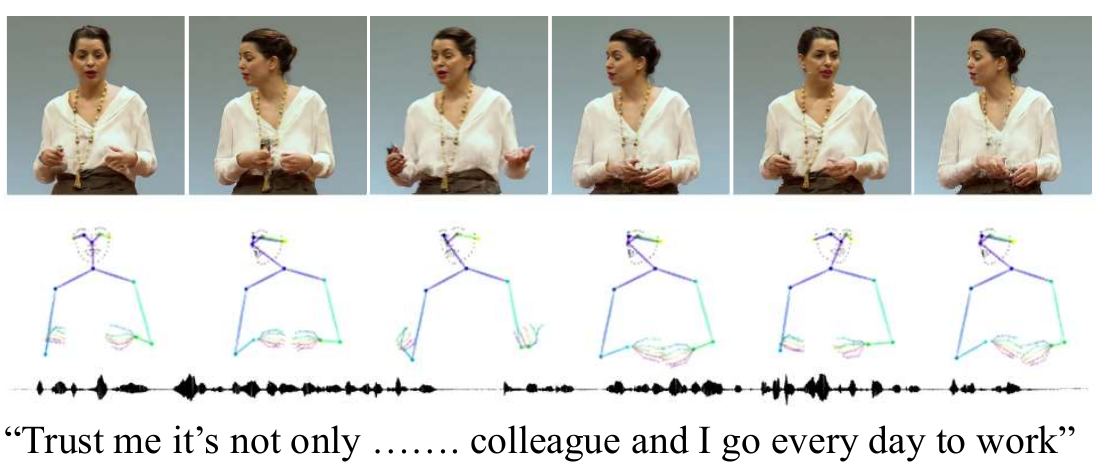}
   \caption{The end-to-end rendering video results. From bottom to top row: speech text, speech audio, generated motion sequence, rendered video frames.}
  \label{fig: e2e}
\end{figure}

We further show the RGB video rendering results based on our generated motions, as depicted in Fig.~\ref{fig: e2e}. We adopt a similar approach as in \cite{everybodydancenow} for video generation. The video results can be end-to-end synthesized given an input speech audio. Interestingly, we can render the generated motions based on arbitrary identities, which enables applications such as generating ``professional" speech videos for amateur speakers by mimicking moves of professional speakers. 

\iffalse
\begin{figure}
    \centering
    \includegraphics[width=\linewidth]{figures/DA.pdf}
    \caption{The data preprocessing and augmentation strategy of our method. For each video segment, we normalize the detected 2D keypoints using the average neck length. During training, we randomly scale the length of each limb to ease the incompatibility of different detected body shapes.}
    \label{fig:DA}
\end{figure}
\fi

%%%%%%%%% BODY TEXT
\section{Conclusion and Discussion}
In this work, we propose the FreeMo for co-speech motion generation. Motivated by studies in linguistics and psychology, we address the non-deterministic mapping by decomposing the hard task into complementary parts. Given an input speech audio, the pose mode branch sequentially generates diverse pose modes with conditional sampling, and the rhythmic motion branch concurrently enriches each pose mode with audio-conditional rhythmic dynamics for spontaneous motions. Our model generates highly diverse and visually plausible body motions in a freeform manner. %Future directions may include incorporating speech semantics, or generating listening behavior. 

\noindent\textbf{Open Issues.}
Our method is based on 2D skeleton landmarks that lack kinetics constrains. Thus current form is sensitive to severe body shape deformations (i.e., bone length). Possible solutions may include adopting a 3D model to represent body and its motion ($M$). 
%Different speakers or camera positions lead to different detected body shapes, which causes incompatibility of the data in different speech videos. We rely on data pre-processing and augmentation as depicted in Fig.~\ref{fig: DA} to ease this problem. However, the diversity and quality are still heavily affected by this problem. 
Moreover, as we do not consider full spectrum of semantic postures, our model is limited in handling the substantial gestures \cite{gesture-visible_action_as_utterance}) that convey conventionalized semantic meaning (e.g., greetings). A sound solution is to incorporate pre-defined semantic gestures \cite{speech2video_synthesis} into the pipeline.

\noindent\textbf{Broader impact.}
This work helps build virtual agents to relieve human labor, which is essential for applications such as social robots adopted in Robotics and digital avatars prevalent in Metaverse. Also, there is possibility towards abused fake video generation. Right now, the negative impact is limited, since this technique is far from mature, not to mention the extensive works on identifying computer generated contents \cite{face_x_ray, magdr, multi_attentional}. 
\newpage

%Discussion of potential negative societal impact: As the AI community at large has been paying increasing attention to the issue of negative societal impact of research, CVPR 2022 aims to raise awareness of potential negative societal impact in the CVPR community as well, as described in the ethics guidelines. To that end, we encourage all authors to think about this issue in the context of the technologies they developed and discuss that in their paper. Note that there is no formal requirement to include a discussion of potential negative societal impact, but reviewers will be asked to favorably consider the inclusion of a meaningful discussion of this issue.

%%%%%%%%% REFERENCES
{\small
\bibliographystyle{ieee_fullname}
\bibliography{egbib}
}

\end{document}